\documentclass[11pt,a4paper]{article}
\usepackage{times,latexsym}
\usepackage{url}
\usepackage[T1]{fontenc}
\usepackage{graphicx}
\usepackage{csquotes}
\usepackage{algorithm}
\usepackage{algpseudocode}
\usepackage{multirow}
\usepackage{booktabs}
\usepackage{microtype}
\usepackage{colortbl}
\usepackage[table,x11names,dvipsnames]{xcolor}
\usepackage[acceptedWithA]{tacl2021v1}
\newcommand{\ci}[2]{{\scriptsize(#1{\tiny$\pm$}#2)}}
\newcommand{\eob}{\texttt{<eob>}}
\newcommand{\eol}{\texttt{<eol>}}

\title{Direct Speech Translation for Automatic Subtitling}

\author{
  Sara Papi\textsuperscript{1,2}, Marco Gaido\textsuperscript{1,2}, Alina Karakanta\textsuperscript{3,}\Thanks{Work done while in FBK.}, \\ \textbf{Mauro Cettolo\textsuperscript{1}, Matteo Negri\textsuperscript{1}, Marco Turchi\textsuperscript{4,*}
  }
  \\
  \textsuperscript{1}Fondazione Bruno Kessler, Trento, Italy
  \\
  \textsuperscript{2}University of Trento, Trento, Italy
  \\
  \textsuperscript{3}Leiden University Centre for Linguistics, Leiden, The Netherlands
  \\
  \textsuperscript{4}Zoom Video Communications, Karlsruhe, Germany
  \\
  \texttt{\{spapi,mgaido,cettolo,negri\}@fbk.eu}
}

\date{}

\begin{document}
\maketitle
\begin{abstract}
Automatic subtitling is the task of automatically translating the speech of audiovisual content into short pieces of timed text, i.e. subtitles and their corresponding timestamps. The generated subtitles need to conform to space and time requirements, while being synchronised with the speech and segmented in a way that facilitates comprehension. Given its considerable complexity, the task has so far been addressed through a pipeline of components that separately deal with transcribing, translating, and segmenting text into subtitles, as well as predicting timestamps. In this paper, we propose the first direct ST model for automatic subtitling that generates subtitles in the target language along with their timestamps with a single model.
Our experiments on 7 language pairs show that our approach outperforms a cascade system in the same data condition, also being competitive with production tools on both in-domain and newly-released out-domain benchmarks covering new scenarios.
\end{abstract}

\section{Introduction}

With the growth of websites and streaming platforms such as YouTube and Netflix,\footnote{\url{https://www.insiderintelligence.com/insights/ott-video-streaming-services/}} the amount of audiovisual content available online has dramatically increased. Suffice to say that the number of hours of Netflix original content has increased by 2,400\% from 2014 to 2019.\footnote{\url{https://www.statista.com/statistics/882490/netflix-original-content-hours/}}
This phenomenon has led to a huge demand for subtitles, which is becoming more and more difficult to satisfy only with human resources.
Consequently, automatic subtitling tools are spreading to reduce subtitlers' workload by providing them with suggested subtitles to be post-edited \citep{alvarez,Vitikainen_Koponen_2021}.
In general, subtitles can be either \textit{intralingual} (hereinafter \textit{captions}), if source audio and subtitle text are in the same language, or \textit{interlingual} (hereinafter \textit{subtitles}), if the text is in a different language. 
In this paper, we focus on automatizing interlingual subtitling, framing it as a speech translation (ST) for subtitling problem.

Differently from ST, in automatic subtitling the generated text has to comply with multiple requirements related to its length, format, and the time it should be displayed on the screen \citep{cintas-ramael-21-sub}. 
These requirements, which depend on the type of video content and target language, are dictated by the need to keep users' cognitive effort as low as possible while maximizing comprehension and engagement \citep{perego-2008-break,szarkowska-2018-fastsubs}. 
This often leads to a condensation of the original spoken content, aimed at reducing the time required for reading subtitles while increasing that of watching the video \citep{Burnham-etal-2008-reduction,Szarkowska-etal-2016-editing}.

Being such a complex task, automatic subtitling has so far been addressed by dividing the process into different steps \citep{piperidis-etal-2004-multimodal,melero-2006-etitle,matusov-etal-2019-customizing,koponen-etal-2020-mt,bojar-etal-2021-elitr}: automatic speech recognition (ASR), timestamp extraction from audio, segmentation into captions, and their machine translation (MT) into the final subtitles. 
More recently, drawing from the evidence that direct models achieve competitive quality with cascade architectures \citep{ansari-etal-2020-findings}, \citet{karakanta-etal-2020-42} proposed an ST system that jointly translates and segments into subtitles, arguing that direct models are able to better exploit speech cues and prosody in subtitle segmentation. However, their system does not generate timestamps, hence missing a critical aspect to reach the goal of fully automatic subtitling. 
Furthermore, the current lack of benchmarks hinders a thorough evaluation of the technologies developed for automatic subtitling.
In fact, the only corpus publicly available to date is MuST-Cinema \citep{karakanta-etal-2020-must}, which contains only single-speaker audios in the TED-talks domain with verbatim translations.

To fill these gaps, this paper presents the first automatic subtitling system that performs the whole task with a single direct ST model, and introduces two new benchmarks.
Our contributions can be summarized as follows:
\begin{itemize}
    \item We propose the first direct ST model for automatic subtitling able to produce both subtitles and timestamps.
    Code and pre-trained models are released under the Apache License 2.0 at: \url{https://github.com/hlt-mt/FBK-fairseq/};
    \item We introduce two (en$\rightarrow$\{de, es\}) benchmarks for automatic subtitling, covering new domains, news/documentaries and interviews, with the presence of background noise and multiple speakers. We release them under the CC BY-NC 4.0 license at: \url{https://mt.fbk.eu/ec-short-clips/} and \url{https://mt.fbk.eu/europarl-interviews/};
    \item We conduct the first extensive comparison between automatic subtitling systems based on cascade and direct ST models on all the 7 language pairs of MuST-Cinema (en$\rightarrow$\{de,es,fr,it,nl,pt,ro\}), showing the superiority of our direct solution, while also demonstrating its competitiveness with production systems on both MuST-Cinema and out-of-domain benchmarks.
\end{itemize}

\section{Background}

\subsection{Direct Speech Translation}
\label{subsec:direct}
While the first cascaded approach to ST was proposed decades ago \citep{cascade,cascade2}, direct models\footnote{According to the official IWSLT definition (\url{https://iwslt.org/2023/offline}), a direct model is a system that does not use intermediate discrete representations to generate the outputs from audio segments and whose parameters used during decoding are all trained altogether on the ST task, while it does not consider the audio segmentation.} have recently become increasingly popular \citep{berard_2016,weiss2017sequence} due to their ability to avoid error propagation \citep{sperber-paulik-2020-speech}, their superior exploitation of prosody and better audio comprehension \citep{bentivogli-etal-2021-cascade}, and their lower computational cost \citep{weller-etal-2021-streaming}.
Motivated by these advantages, direct models are rapidly evolving and their initial performance gap with cascade architectures \citep{niehues-etal-2019-iwslt} has been significantly reduced, leading to a substantial parity in the latest IWSLT campaigns \citep{ansari-etal-2020-findings,anastasopoulos-etal-2021-findings,anastasopoulos-etal-2022-findings}.
Such improvements can be partly attributed to the development of specialized architectures for speech processing \citep{Chang2020EndtoEndAW,papi-etal-2021-speechformer,efficient-conformer,kim2022squeezeformer,andrusenko2022uconv}, which are all variants of a Transformer model \citep{transformer} preceded by convolutional layers that reduce the length of the input sequence \citep{berard_2018,gangi19_interspeech}.
Among them, Conformer \citep{gulati20_interspeech} is currently the best-performing model in ST \citep{inaguma2021non}. For this reason, we build our systems with this architecture and test, for the first time, its effectiveness in the challenging task of fully automatic subtitling.

\subsection{Subtitling Requirements}
\label{sec:guidelines}
Subtitles are short pieces of timed text, generally displayed at the bottom of the screen, which describe, transcribe, or translate the dialogue or narrative. A subtitle is composed of two elements: the text, shown into ``blocks'', and the corresponding start and end display time -- or timestamps.\footnote{The most widespread subtitle format is SubRip or srt.}

Depending on the subtitle provider and the audiovisual content, different requirements have to be respected concerning both the text space and its timing.
These constraints typically consist in: \textit{i)} using at most two lines per block; \textit{ii)} keeping linguistic units (e.g. noun and verb phrases) in the same line; \textit{iii)} not exceeding a pre-defined number of characters per line (CPL), spaces included; \textit{iv)}  not exceeding a pre-defined reading speed, measured in number of characters per second (CPS).
While a typical value used as maximum CPL threshold is 42 for most Latin languages,\footnote{\url{https://www.ted.com/participate/translate/subtitling-tips}} there is no agreement on the maximum CPS allowed. For instance, Netflix guidelines\footnote{ \url{https://partnerhelp.netflixstudios.com/hc/en-us/articles/219375728-Timed-Text-Style-Guide-Subtitle-Templates}} allow up to 17 CPS for adult and 15 for children programs, TED guidelines\footnote{\url{https://www.ted.com/participate/translate/subtitling-tips}} up to 21 CPS, 
and Amara guidelines\footnote{\url{https://blog.amara.org/2020/10/22/create-quality-subtitles-in-a-few-simple-steps/}} up to 25 CPS.

To convey the meaning of the audiovisual product while adhering to time and space constraints, in some domains and scenarios subtitles require compression or condensation \citep{doi:10.1080/0907676X.2001.9961416,condensation,aziz-etal-2012-cross,liu-etal-2020-adapting,buet21_interspeech}. 
Due to the rehearsed nature of TED talks, the subtitles in MuST-Cinema have a limited degree of condensation, and the translation is mostly verbatim.
In addition, the audio conditions (no background noise and a single speaker) are not representative of all the diverse contexts where subtitling is applied, such as news and movies. To fill this gap, we introduce two new benchmarks that feature different domains, scenarios (e.g., multiple speakers), and levels of subtitle condensation.

\subsection{Automatic Subtitling}
\label{subsec:automatic-subtitling}
Attempts to (semi-)automatize the subtitling process have been done with cascade systems made of an ASR, a segmenter, and an MT model.
Most works focused on adapting the MT module to subtitling with the goal of producing shorter and compressed texts.
This has been performed either using statistical approaches trained on subtitling corpora \citep{volk-etal-2010-machine,etchegoyhen-etal-2014-machine,smtsubtitling} or by developing specifically tailored decoding solutions on statistical \citep{aziz-etal-2012-cross} and neural models \citep{matusov-etal-2019-customizing}. 
In particular, recent research efforts focused on controlling the MT output length so as to satisfy isometric requirements between source transcripts and target translations \citep{lakew-etal-2019-controlling,matusov2020flexible,isometric2021,isometric2022}.
In addition, \citep{ktem2019ProsodicPA,Federico2020,9414966,tam22_interspeech,Effendi2022DurationMO} proved the usefulness of injecting prosody information about speech cues, such as pauses, in determining subtitle boundaries.
Given the possibility for direct ST systems to access this information and their advantages mentioned in \S\ref{subsec:direct}, \citet{karakanta-etal-2020-42,karakanta-etal-2021-flexibility} built the only (to the best of our knowledge) automatic subtitling system using a direct ST model, confirming with their results that the ability of direct ST systems to leverage prosody has particular importance for subtitle segmentation. However, their solution only covers the translation and segmentation into subtitles, neglecting the timestamp generation. 
Our study is hence the first to complete the entire subtitling process with a direct ST model and to evaluate its performance on all aspects of the subtitling task.

\begin{figure*}[!t]
    \centering
    \includegraphics[width=\textwidth]{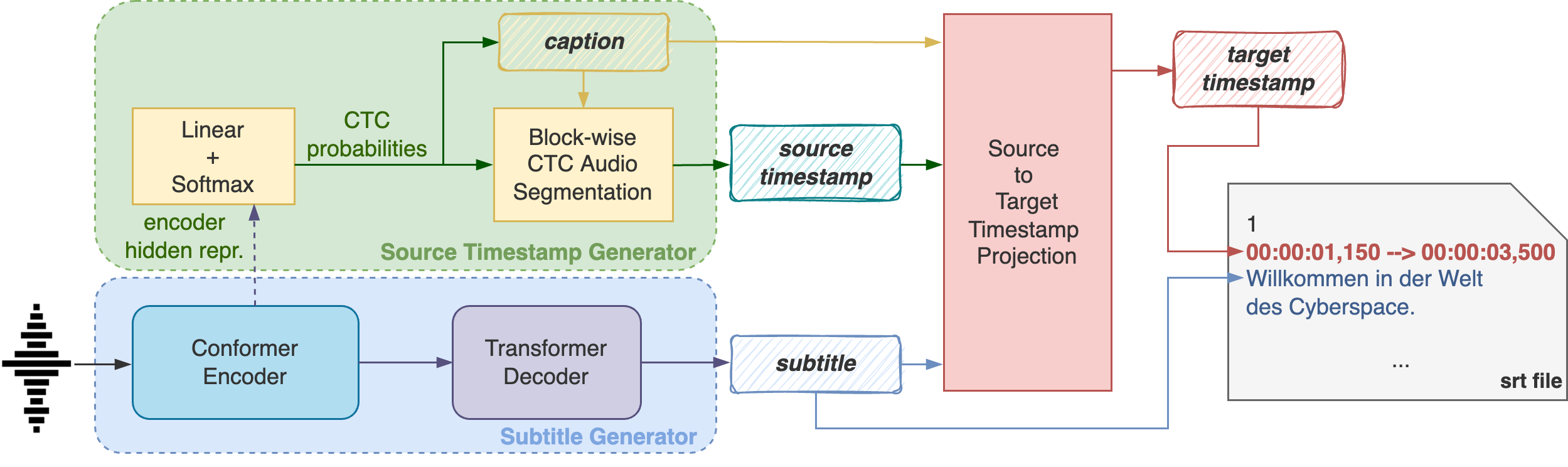}
    \caption{Architecture of the direct ST system for automatic subtitling.}
    \label{fig:model}
\end{figure*}

\section{Direct Speech Translation for Subtitling}
\label{sec:model}

Motivated by all the advantages discussed in \S\ref{subsec:direct} and \S\ref{subsec:automatic-subtitling}, we build the first automatic subtitling system solely based on a direct ST model (Figure~\ref{fig:model}).
Our system works as follows: \textit{i)} the audio is fed to a \textit{Subtitle Generator}~(§\ref{subsec:textualout}) that produces the (untimed) subtitle blocks; \textit{ii)}~the computed encoder representations are passed to the \textit{Source Timestamp Generator} (§\ref{subsec:CTCsegm}) to obtain the caption blocks and their corresponding timestamps; \textit{iii)} the subtitle timestamps are estimated by the \textit{Source-to-Target Timestamp Projection} (§\ref{subsec:timestamp}) from the generated subtitles, captions, and source timestamps. 
These modules are described in the rest of this section.

\subsection{Subtitle Generation}
\label{subsec:textualout}

We train a direct ST Conformer-based model that jointly performs the ST task and the segmentation of the generated translation into (untimed) subtitle blocks and lines. 
To this end, we add two special tokens to the vocabulary of our system, \eob~and \eol, which respectively represent the end of a subtitle block and the end of a line within a block. 
Both at training and inference time, \eob~and \eol~are treated as any other token, without giving them different weights, or adding specific loss.
Additionally,  we do not incorporate losses aimed at minimizing the number of generated characters or explicitly optimizing for CPL and CPS compliance.

\subsection{Source Timestamp Generation}
\label{subsec:CTCsegm}

Estimating timestamps for the generated subtitle blocks from source audio is a challenging task.
Current sequence-to-sequence models, in fact, generate target sequences that are decoupled from the input and, therefore, their tokens do not have a clear relationship with the frames they correspond to.
To recover this relationship, we start from the observation that direct ST models are often trained with an auxiliary Connectionist Temporal Classification or CTC loss \citep{ctc-2006} in the encoder to improve model convergence \cite{kim-et-al-2017-joint,bahar2019comparative}. The CTC maps the input frames to the transcripts~--~in our use case, captions~--~ and we propose to leverage this CTC module at inference time to estimate the block timestamps.

In particular, the encoder representations computed during the forward pass are fed to the CTC module that provides the frame-level probability distribution over the source vocabulary tokens (including \eob, \eol, and the additional CTC \textit{blank} token). 
This sequence of CTC probabilities over the source vocabulary serves two purposes.
First, it is used to predict the caption with the CTC beam search algorithm \citep{graves_ctc_beam_search}.\footnote{We also tested a greedy decoding, in which the most likely label for each time step is chosen to obtain the output sequence. However, this approach did not prove effective.}
Second, it is fed, together with the generated caption, to the CTC-based segmentation algorithm \citep{ctcsegmentation}, whose task is to find the most likely alignment between caption tokens and audio frames.
The algorithm builds a trellis over the time steps for the generated tokens and, at each time step, only three paths are possible: \textit{i)} staying at the same token (self-loop); \textit{ii)} moving to the \textit{blank} token; \textit{iii)} moving to the next token. 
To avoid forcing the caption to start at the beginning of the audio, the transition cost for staying at the first token is set to 0.
Otherwise, the transition cost is the CTC-predicted probability for a given token in that time step.
The trellis is then backtracked from the time step with the highest probability in the last token of the generated caption, until the first token is reached.
In our case, since we are interested in the timestamps of the subtitle blocks, we extract block-wise alignments that correspond to the start and the end time of each block. This means finding the time in which the first word of each subtitle is pronounced and the time in which the corresponding \eob~symbol is emitted by using the aforementioned algorithm.

\subsection{Source-to-Target Timestamp Projection}
\label{subsec:timestamp}

After generating the untimed subtitles (§\ref{subsec:textualout}), and captions with their timestamps (§\ref{subsec:CTCsegm}), the next step is to obtain the timestamps for subtitle blocks on the target side.
In general, caption and subtitle segmentations may differ for many reasons (e.g. due to different syntactic patterns between languages) and imposing the caption segmentation on the subtitle side -- as done in most cascade approaches \citep{Georgakopoulou_2019,koponen-etal-2020-mt} -- could be a sub-optimal solution.
For this reason, we introduce a caption-subtitle alignment module that projects the source timestamps to the target blocks. 
To perform this task, we tested the three alternative methods described below.

\begin{figure}[!tb]
    \centering
    \includegraphics[width=0.43\textwidth]{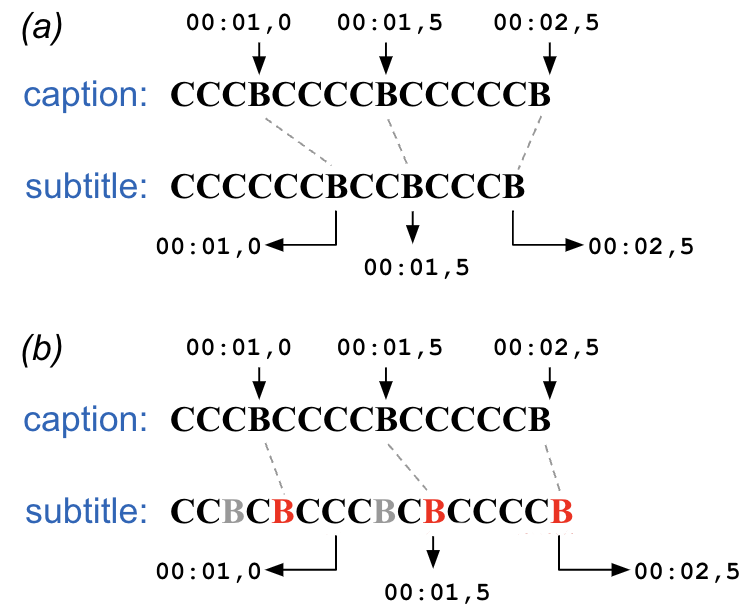}
    \caption{Example of BWP projection with \textit{(a)} same number of blocks and \textit{(b)} different number of blocks between caption and subtitle.}
    \label{fig:bwp}
\end{figure}

\paragraph{Block-Wise Projection (BWP)}
This method operates at character level to project the predicted source-side (captions) timestamps on the target side (subtitles) without alterations.
When the number of caption and subtitle blocks is equal, a condition that occurs in $\sim$80\% of the cases, the timestamps of each caption block are directly assigned to the corresponding subtitle block.\footnote{Selecting the candidates with the closest number of blocks among the source and target $n$-best lists had negligible effects.} This process is depicted in Figure \ref{fig:bwp}.a, in which ``C'' and ``B'' respectively stand for characters and blocks in the caption and subtitle.
When the number of caption and subtitle blocks is different (Figure \ref{fig:bwp}.b), the target segmentation is discarded and replaced with the caption segmentation. In this case, line and block boundaries (\eol/\eob) are inserted in the target side by matching the number of characters each line/block has in the caption. 
If the insertion falls in the middle of a word, the \eol/\eob~is appended to the word.
This approach has two main weaknesses. First, it assumes that, when captions and subtitles have the same number of blocks, these blocks contain the same linguistic content, while this is not guaranteed. Second, it ignores the subtitle segmentation in $\sim$20\% of the cases.

\begin{figure}[!htb]
    \centering
    \includegraphics[width=0.43\textwidth]{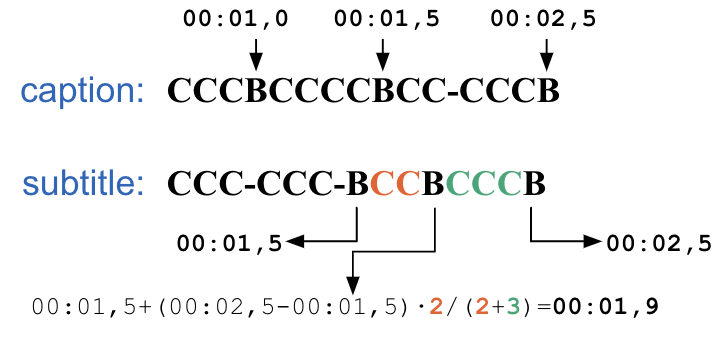}
    \caption{Example of Levenshtein-based projection.}
    \label{fig:alignment}
\end{figure}

\paragraph{Levenshtein-based Projection (LEV)}
To overcome the above limitations, our second method exploits the Levenshtein distance-based alignment \citep{Levenshtein_SPD66} between captions and subtitles.
This method estimates the target-side timestamps from the source-side timestamps without ever altering the original target-side segmentation.
First, all the non-block characters are masked with a single symbol (\enquote{C}). For instance, \enquote{\emph{This is a block} \eob} is converted into \enquote{CCCCCCCCCCCCCCCB}, where \enquote{B} stands for \eob.
Then, the masked caption and subtitle are aligned with the weighted version of Levenshtein distance, in which the substitution operation is forbidden so as to avoid the replacement of a character with a block and vice versa.
If the positions of a block in the aligned caption and subtitle match, its caption timestamp is directly assigned to the subtitle block.
If they do not match, the timestamps of the subtitle blocks are estimated from the caption timestamps based on the alignment of ``B''s and the number of characters.
For instance, given the caption \enquote{CCCBCCCCBCCCCCB} and the subtitle \enquote{CCCCCCBCCBCCCB}, the optimal source-target alignment with the corresponding timestamp calculation is shown in Figure \ref{fig:alignment}.
In detail, the first subtitle block (CCC-CCC-B) is matched with the first two caption blocks (CCCBCCCCB) and the corresponding timestamp (\texttt{00:01,5}) is directly mapped. This also happens with the timestamp \texttt{00:02,5} of the last caption (BCC-CCCB) and subtitle block (\textcolor{teal}{CCC}B).
For the second subtitle block (\textcolor{orange}{CC}B), the timestamp (\texttt{00:01,9}) is estimated proportionally from the caption (BCC-CCCB) using the character ratio between the orange block and the orange + green blocks. 

\begin{figure}[!htb]
    \centering
    \includegraphics[width=0.47\textwidth]{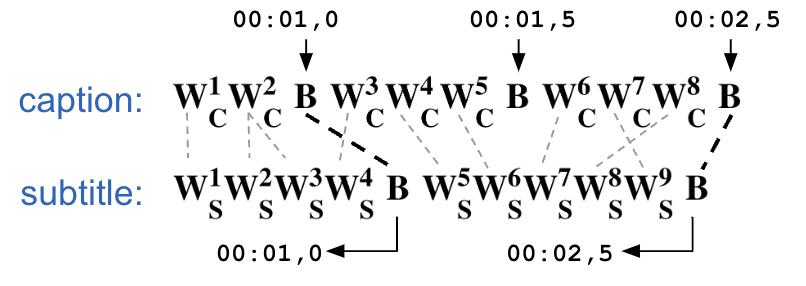}
    \caption{Example of Semantic-based projection.}
    \label{fig:sem}
\end{figure}

\paragraph{Semantic-based Projection (SEM)}
The third method projects the predicted source-side timestamps on target blocks by looking at the semantic content of the generated captions and subtitles. The method is based on SimAlign \citep{jalili-sabet-etal-2020-simalign}, which combines semantic embeddings from fastText \citep{10.1162/tacl_a_00051}, VecMap \citep{artetxe-etal-2018-robust}, mBERT,\footnote{\url{https://github.com/google-research/bert/blob/master/multilingual.md}} and XLM-RoBERTa \citep{conneau-etal-2020-unsupervised} to align source and target texts at the word level.
Specifically, we first align captions and subtitles word by word (\eol{}/\eob{} included) with SimAlign.
Then, when all \eob{}s of a subtitle are aligned with \eob{}s in the caption (66\% of the cases), we assign the corresponding timestamp (Figure \ref{fig:sem}). Otherwise, i.e. when at least one \eob{} in the subtitle is aligned with a caption word or \eol{} or is not aligned at all, one of the two previous methods is applied as a fallback solution.

\section{Data}
\label{subsec:data}

\subsection{Training Data}
\label{subsec:train-data}
For the comparison between cascade and direct architectures (\S\ref{subsec:direct_vs_cascade}), we train the models in a controlled and easily reproducible data setting by using MuST-Cinema v1.1, the only publicly available subtitling corpus also containing the source speech. 
It covers one general domain (TED talks), and 7 language pairs, namely en$\rightarrow$\{de, es, fr, it, nl, pt, ro\}. 
The number of hours in the training set of each language pair is shown in the first row of Table \ref{tab:data_hours}.

\begin{table}[tb]
\small
    \centering
    \setlength{\tabcolsep}{3pt}
    \begin{tabular}{lccccccc}
    \specialrule{.1em}{.05em}{.05em} 
        \textbf{Dataset} & \textbf{de} & \textbf{es} & \textbf{fr} & \textbf{it} & \textbf{nl} & \textbf{pt} & \textbf{ro} \\
        \hline
        MuST-Cinema & 388 & 479 & 469 & 441 & 421 & 364 & 410 \\
        Europarl-ST & 75 & 74 & - & - & - & - & - \\
        CoVoST2 & 412 & 412 & - & - & - & - & - \\
        CommonVoice & 885 & 885 & - & - & - & - & - \\
        TEDlium & 444 & 444 & - & - & - & - & - \\
        VoxPopuli & 519 & 519 & - & - & - & - & - \\
    \specialrule{.1em}{.05em}{.05em} 
    \end{tabular}
    \caption{Number of hours of the training sets.}
    \label{tab:data_hours}
\end{table}

For the comparison with production tools (\S\ref{subsec:production-results}), we experiment in a more realistic unconstrained data scenario and we focus on en$\rightarrow$de, and en$\rightarrow$es.\footnote{We select these two language pairs due to, respectively, a different and similar word ordering with respect to the source.} 
For training, we use MuST-Cinema, two ST datasets -- Europarl-ST \citep{europarlst} and CoVoST2 \citep{wang2020covost} -- and three ASR datasets --
CommonVoice \citep{ardila-etal-2020-common}, TEDlium \cite{tedlium} and VoxPopuli \citep{wang-etal-2021-voxpopuli}.
We translate the ASR corpora with the Helsinki-NLP MT models \citep{TiedemannThottingal:EAMT2020} and filter out data with a very high
or low transcript/translation character ratio, as per \citep{gaido-etal-2022-efficient}.
The use of automatic translations as targets, also known as sequence-level knowledge distillation \citep{kim-rush-2016-sequence}, is a popular data augmentation method used in the most recent IWSLT evaluation campaigns \citep{anastasopoulos-etal-2021-findings,anastasopoulos-etal-2022-findings} to enhance the performance of ST systems. 
Since none of the training sets, except for MuST-Cinema, includes the subtitle boundaries (\eob~and \eol) in the target translation, we automatically insert them by employing the publicly-released multimodal and multilingual segmenter by \citet{segmenter}.
The segmenter takes the source audio and the unsegmented text as input and outputs the segmented text i.e., containing \eob~and \eol. 
By doing this, we can train our system to jointly translate from speech and segment into subtitles without the need for manually-curated subtitle targets, which are hard to find and costly to create.
The number of training hours is reported in Table \ref{tab:data_hours}.

\subsection{Test Data}
The models are tested in both in-domain and out-of-domain conditions.
For in-domain experiments, we use the MuST-Cinema test set, for which we adopt both the original audio segmentation (for reproducibility and for the sake of comparison with previous and future work) and more realistic automatic segmentation obtained with SHAS \citep{tsiamas2022shas}.
Notice that this audio segmentation is a completely different task from determining subtitle boundaries. Its only goal is splitting long audio files into smaller chunks (or utterances) that can be processed by ST systems, limiting performance degradation due to information loss caused by sub-optimal splits (e.g., in the middle of a sentence). In general, each resulting utterance contains multiple subtitle blocks.
For instance, in the MuST-Cinema training set there are $\sim$2.5 blocks per utterance, even though utterances are quite short (6.4s on average). When automatic segmentation methods like SHAS are applied, this ratio significantly increases, as audio segments are typically much longer, with many segments lasting between 14 and 20 seconds \citep{gaido-etal-2021-beyond,tsiamas2022shas}.

For out-of-domain evaluations, we introduce the two new (en$\rightarrow$\{de,es\})
test sets described below, which we also segment with SHAS.

\paragraph{EC Short Clips}
The first test set is composed of short videos from the Audiovisual Service of the European Commission (EC)\footnote{\url{https://audiovisual.ec.europa.eu/}} recorded between 2016 and 2022. These informative
clips have an average duration of 2 minutes and cover various topics discussed in EC debates such as economy, environment, and international rights.
This benchmark
presents several additional difficulties compared to TED talks since the videos often contain multiple speakers, and background music is sometimes present during the speech.
We selected the videos with the highest subtitle conformity (at least 80\% of the subtitles conforming to 42 CPL, and 75\% conforming to 21 CPS), and removed subtitles describing on-screen text. This resulted in 27 videos having a total duration of 1 hour. The target srt files contain $\sim$5,000 words per language.

\paragraph{EuroParl Interviews} The second test set is compiled from publicly available video interviews from the European Parliament TV\footnote{\url{https://www.europarltv.europa.eu/}} (2009-2015). We selected 12 videos of 1 hour total duration, amounting to $\sim$6,500 words per target language.
The videos present multiple speakers and sometimes contain short interposed clips with news or narratives. 
Apart from the more challenging source audio properties compared to the clean single-speaker TED talks, here the target subtitles are not verbatim and demonstrate a high degree of compression and reduction. 
As a consequence, the CPL and CPS conformity is very high ($\sim$100\%) but this comes at the cost of being more difficult for automatic systems to perfectly match the non-verbatim translations. Nonetheless, to achieve real progress in automatic subtitling, it is particularly relevant to evaluate automatic systems on realistic and challenging benchmarks like the ones we provide. 

\section{Experimental Settings}

\subsection{Training Settings}
Our systems are implemented on Fairseq-ST \cite{wang2020fairseqs2t}, following the default settings unless stated otherwise. 
The input is represented by 80 audio features extracted every 10$ms$ with sample window of 25 and pre-processed by two 1D convolutional layers with stride 2 to reduce the input length by a factor of 4. All segments longer than 30$s$ in the training set are filtered out to speed up training.
The models are based on encoder-decoder architectures and composed by a stack of 12 Conformer encoder layers and 8 Transformer decoder layers. We apply CTC loss to the 8\textsuperscript{th} encoder layer and use its predictions to compress the input sequences to reduce RAM consumption \cite{liu2020bridging,gaido-etal-2021-ctc}.
Both the Conformer and Transformer layers have a 512 embedding dimension and 2,048 hidden units in the linear layer. 
We set dropout to 0.1 in the linear, attention, and convolutional modules.
In the convolutional modules, we also set a kernel size of 31 for the point- and depth-wise convolutions. 

For the comparison between cascade and direct architectures, we train a one-to-many multilingual ST model that prepends a token representing the selected target language for decoding \citep{hirofumi2019_multilingual} on all the 7 languages of MuST-Cinema.
Conversely, for the comparison with production tools, we develop a dedicated ST model for each target language (de, es).
For inference, we set the beam size to 5 for both subtitles and captions.

We train with Adam optimizer \citep{DBLP:journals/corr/KingmaB14} ($\beta_1=0.9$, $\beta_2=0.98$) for 100,000 steps. The learning rate increases linearly up to 0.002 for the first 25,000 warm-up steps and then decays with an inverse square root policy, apart from fine-tunings, where it is the fixed value 0.001.
Utterance-level Cepstral Mean and Variance Normalization (CMVN) and SpecAugment \cite{Park2019} are applied during training, as per Fairseq-ST default settings.
The vocabularies are based on SentencePiece models \citep{sennrich-etal-2016-neural} with size 8,000 for the source language. 
For the multilingual model trained on MuST-Cinema, a shared vocabulary is built with a size of 16,000 while, for the two models developed to compare with production tools, we build German and Spanish vocabularies with a size of 16,000 subwords each.
The ASR of our cascade model is trained using the same source language vocabulary of size 8,000 used in the translation setting.
The MT model is trained using the standard hyper-parameters of the Fairseq multilingual MT task \citep{ott-etal-2019-fairseq}, with the same source and target vocabularies of the ST task. 

For all models, we stop the training when the validation loss does not improve for 10 epochs and the final models are obtained by averaging 7 checkpoints (the best, 3 preceding and 3 succeeding).
Training is performed on 4 NVIDIA A100 (40GB RAM), with 40k max tokens per mini-batch and an update frequency of 2, except for the MT models for which 8 NVIDIA K80 (12GB RAM) are used with 4k max tokens and an update frequency of~1.
Table \ref{tab:params} lists the total number of parameters of our direct models, showing that it is $\sim$1/3 of the cascade system used as a term of comparison.

\subsection{Terms of Comparison} 
\label{subsec:comparison}

We compare our direct ST system both with a cascade pipeline trained under the same data conditions and with production tools.

\begin{table}[!t]
\centering
\small
\begin{tabular}{l|c}
\bottomrule
\textbf{System} & \textbf{Num. params} \\
\hline
\hline
Direct & 124.6M \\
\hline
Cascade & 341.9M \\
\hspace{3mm} - ASR & 116.4M \\
\hspace{3mm} - Audio forced aligner  & 9.7M \\
\hspace{3mm} - Segmenter  & 40.6M \\
\hspace{3mm} - Multilingual MT & 175.2M \\
\hline
\toprule
\end{tabular}
\caption{\label{tab:params} Number of parameters for the direct (both multilingual and monolingual) and cascade systems.}
\end{table}

\paragraph{Cascade}
We build an in-domain cascade composed of: an ASR, an audio forced aligner, a segmenter, and an MT system.
The ASR has the same architecture of our ST system (Conformer encoder + Transformer decoder), and it is trained on MuST-Cinema transcripts without \eob~and \eol. 
The audio forced aligner used to estimate the timestamps \citep{gretter2021seed} is based on the Kaldi\footnote{\url{https://github.com/kaldi-asr/kaldi}} acoustic model.
The subtitle segmenter is the same multimodal segmenter we used to segment the training data for the direct system (§\ref{subsec:train-data}). The MT is a multilingual model trained on the MuST-Cinema (\textit{transcript}, \textit{translation}) pairs without \eob~and \eol.
The pipeline works as follows.
The audio is first transcribed by the ASR and word-level timestamps are estimated with the forced aligner. Then, the transcript is segmented into captions with the segmenter and each block timestamp is obtained by averaging the end time of the word before an \eob{} and the start time of the word after it.
The segmented text is then split into sentences according to the \eob~and, finally, these sentences
are translated by the MT.
The \eob{s} are automatically re-inserted at the end of each sentence while \eol{s} are added to the subtitle translation using the same segmenter.

\paragraph{Production Tools}
As a term of comparison for the unconstrained data condition, we use production tools for automatic subtitling.
These tools take audio or video content as input and return the subtitles in various formats, including srt. 
We test three online tools,\footnote{All outputs were collected in August 2022.} namely: MateSub,\footnote{\url{https://matesub.com/}} Sonix,\footnote{\url{https://sonix.ai}} and Zeemo.\footnote{\url{https://zeemo.ai/}}
We also compare with the AppTek subtitling system,\footnote{\url{https://www.apptek.com/}} a cascade architecture whose ASR component is equipped with a neural model that predicts the subtitle boundaries before feeding the transcripts to the MT component \citep{matusov-etal-2019-customizing}.
For this system, two variants of the MT model are evaluated: a standard model and a model specifically trained to obtain shorter translations in order to better conform to length requirements \citep{matusov2020flexible}. Since we are not interested in comparing the tools with each other, all system scores are anonymized.

\begin{table*}[!t]
\centering
\small
\begin{tabular}{l|cccc|cccc} 
\bottomrule
 & \multicolumn{4}{c|}{\textbf{en-de}} & \multicolumn{4}{c}{\textbf{en-es}}\\
\hline
\rowcolor{gray!10}
Model & SubER ($\downarrow$) & Sigma ($\uparrow$) & CPL ($\uparrow$) & CPS ($\uparrow$) & SubER ($\downarrow$) & Sigma ($\uparrow$) & CPL ($\uparrow$) & CPS ($\uparrow$) \\ 
 \hline
\multicolumn{9}{c}{\textit{Gold audio segmentation}} \\
\hline
Baseline & 63.5 & 65.6 & 77.7 & 64.4 & 52.0 & 70.4 & 80.7 & 68.0 \\
\hline
BWP & 60.8 & 75.6 & 86.1 & 64.0 & 48.6 & 78.5 & 90.9 & 66.9 \\
LEV & \textbf{58.7} & \textbf{78.8}  & \textbf{88.8} & \textbf{65.4} & \textbf{46.7} & \textbf{81.1} & 93.9 & \textbf{68.4} \\
SEM & 60.7 & 75.5 & 88.6 & 63.7 & 48.6 & 78.8 & \textbf{94.0} & 65.5 \\
\hline
\multicolumn{9}{c}{\textit{Automatic audio segmentation}} \\
\hline
Baseline & 66.9 & 62.0 & 78.2 & 70.5 & 55.7 & 66.0 & 79.9 & 75.1 \\
\hline
BWP & 62.8 & 73.3 & 86.2 & 70.3 & 51.8 & 75.9 & 89.6 & 73.5 \\
LEV & \textbf{60.3} & \textbf{78.5} & \textbf{88.9} & \textbf{72.1} & \textbf{48.5} & \textbf{80.6} & \textbf{94.2} & \textbf{76.1}  \\
SEM & 62.8 & 75.8 & \textbf{88.9} & 69.7 & 51.4 & 78.3 & \textbf{94.2} & 72.9 \\
\toprule
\end{tabular}
\caption{Comparison of timestamp projection methods on the MuST-Cinema en$\rightarrow$\{de, es\} test set.}
\label{tab:analysis}
\end{table*}

\subsection{Evaluation}
\label{subsec:eval}
Translation quality, timing, and segmentation of subtitles are measured with multiple metrics.
First, we compute SubER \citep{wilken-etal-2022-suber},\footnote{Version 0.2.0.} a tailored TER-based metric (the lower, the better) that scores the overall subtitle quality by considering translation, segmentation and timing altogether. We adopt the cased and with punctuation version of the metric since these aspects are crucial for the quality and comprehension of the subtitles.
Next, specifically for translation quality, we use SacreBLEU \citep{post-2018-call},\footnote{case:mixed|eff:no|tok:13a|smooth:exp|version:2.3.1} on texts from which \eol~and \eob~have been removed.
The quality of segmentation into subtitles is evaluated with Sigma from the EvalSub toolkit \citep{karakanta2022evaluating}.
Since BLEU and Sigma require the same audio segmentation between reference and predicted subtitles, we re-align the predictions in case of non-perfect alignment with the mWERSegmenter \citep{matusov-etal-2005-evaluating}. 
Lastly, to check the spatio-temporal compliance described in \S\ref{sec:guidelines}, we compute CPL conformity as the percentage of lines not exceeding 42 characters, and CPS conformity as the percentage of subtitle blocks having a maximum reading speed of 21 characters per second.\footnote{We used version 1.1 of the script adopted for the IWSLT subtitling task (\url{https://iwslt.org/2023/subtitling}): \url{https://github.com/hlt-mt/FBK-fairseq/blob/master/examples/speech_to_text/scripts/subtitle_compliance.py}.}
Confidence intervals (CI) are computed with bootstrap resampling \cite{koehn-2004-statistical}.

\section{Results}
In this section, we first (\S\ref{subsec:timestampexp}) choose the best timestamp projection method among those introduced in \S\ref{subsec:timestamp}. Then (\S\ref{subsec:direct_vs_cascade}), we compare the cascade and direct approaches trained in the same data conditions. Lastly (\S\ref{subsec:production-results}), we show that our direct model, even though trained in laboratory settings, is competitive with production tools.
In addition, in Appendix \ref{sec:force-align}, we analyze the performance of the CTC-segmentation algorithm for timestamp estimation compared to forced aligner tools.

\subsection{Timestamp Projection}
\label{subsec:timestampexp}

The quality of source-to-target timestamp projection (§\ref{sec:model}) is crucial to correctly estimate the target-side timestamps and, in turn, to produce good subtitles.
To select the best strategy, we compare the methods in §\ref{subsec:timestamp} using the constrained model on the MuST-Cinema test sets for en$\rightarrow$\{de, es\}.
To test the robustness of the various methods when gold-segmented audio is not available, we also report the results using the automatic audio segmentation in addition to that obtained using the gold one.

Results are shown in Table \ref{tab:analysis}.
BLEU~is not reported because the translated text is always the same, regardless of the timestamp projection method.
We also report, as a baseline, a method that completely ignores the target segmentation and always maps the caption segmentation onto the subtitle as in BWP when the number of caption and subtitle blocks is different (\S\ref{subsec:timestamp}).
For the SEM method, if the source-target alignment is not found by SimAlign, the LEV method is applied instead.\footnote{We also applied the baseline and the BWP method as a fallback method for SEM but it led to worse results.}

The results highlight the superiority of the LEV method, which outperforms the others on almost all metrics, with similar trends for both language pairs.
The gap is more marked in the realistic scenario of automatically-segmented audio, likely due to the fact that the audio segments produced by SHAS are longer than the manually-annotated ones (8.6s vs 5.5s).
As such, each audio segment contains more blocks to align, so the difference between the methods emerges more clearly. 
The low scores obtained by the baseline confirm that the caption segmentation is not optimal for the target language.
Furthermore, SEM yields results that are either comparable to or slightly better than those obtained by BWP, especially in terms of Sigma and CPL, while being always worse than LEV.
In addition, SEM exhibits lower CPS conformity even compared to the baseline.
Consequently, its performance suggests that semantically-motivated approaches are not the best solution for timestamp projection.

Focusing on the LEV method, we observe that segmentation quality (higher Sigma) and overall subtitle quality (lower SubER) are slightly better when the gold segmentation is used, as expected. Conversely, CPS conformity is higher with the automatic audio segmentation.
This counter-intuitive result can be explained as follows: audio segmentation not only splits but sometimes also cuts the audio according to speakers' pauses, while the manual segmentation delimits speech boundaries more aggressively than the automatic one. In our case, manual segmentation results in audio segments that are about 2\% shorter than those obtained with the automatic segmentation, thus \enquote{forcing} the generated subtitles to appear on screen for a shorter time, which in turn leads to a higher reading speed.

\begin{table*}[tb]
\centering
\setlength{\tabcolsep}{2.5pt}
\small
\resizebox{2.08\columnwidth}{!}{%
\begin{tabular}{l|ccccccc|c} 
\bottomrule
\textbf{Sys.} & \textbf{en-de} & \textbf{en-es} & \textbf{en-fr} & \textbf{en-it} & \textbf{en-nl} & \textbf{en-pt} & \textbf{en-ro} & \textbf{Avg.}\\ 
\hline
& \multicolumn{8}{c}{\cellcolor{white} \textbf{SubER ($\downarrow$)}} \\
\hline
\rowcolor{gray!20} Casc. & 64.2 \ci{64.2}{2.4} & 50.5 \ci{50.5}{2.2} & 57.0 \ci{57.0}{1.8} & 54.2 \ci{54.2}{1.7} & 52.8 \ci{52.8}{1.8} & 49.7 \ci{49.7}{1.7} & 52.7 \ci{52.7}{2.0} & 54.4 \\ 
Dir. & \textbf{58.7} \ci{58.7}{2.3} & \textbf{46.7} \ci{46.7}{2.1} & \textbf{52.9} \ci{52.9}{1.7} & \textbf{50.4} \ci{50.4}{1.7} & \textbf{47.4} \ci{47.4}{1.9} & \textbf{44.6} \ci{44.6}{1.7} & \textbf{48.5} \ci{48.5}{2.1} & \textbf{49.9} \\ 
\hline
& \multicolumn{8}{c}{\cellcolor{white} \textbf{BLEU ($\uparrow$)}} \\
\hline
\rowcolor{gray!20}  Casc. & 18.9 \ci{18.9}{1.4} & 32.4 \ci{32.4}{1.8} & 25.1 \ci{25.1}{1.5} & 26.0 \ci{26.0}{1.6} & 25.8 \ci{25.8}{1.5} & 31.4 \ci{31.4}{1.7} & 28.4 \ci{28.3}{1.6} & 26.9 \\ 
Dir. & \textbf{22.1} \ci{22.1}{1.6} & \textbf{35.9} \ci{35.8}{1.9} & \textbf{28.0} \ci{28.0}{1.6} & \textbf{29.6} \ci{29.6}{1.8} & \textbf{31.6} \ci{31.6}{1.8} & \textbf{36.8} \ci{36.7}{1.7} & \textbf{31.9} \ci{31.8}{1.8} & \textbf{30.8} \\ 
\hline
& \multicolumn{8}{c}{\cellcolor{white} \textbf{Sigma ($\uparrow$)}} \\
\hline
\rowcolor{gray!20}  Casc. & \textbf{79.5} \ci{79.5}{2.0} & 80.9 \ci{80.9}{1.5} & 84.0 \ci{84.0}{1.7} & 83.8 \ci{83.8}{1.6} & 77.5 \ci{77.4}{1.8} & 81.2 \ci{81.2}{1.7} & \textbf{86.4} \ci{86.4}{1.5} & 81.9 \\ 
Dir. & 78.8 \ci{78.8}{2.0} & \textbf{81.1} \ci{81.1}{1.5} & \textbf{84.1} \ci{84.1}{1.7} & \textbf{85.1} \ci{85.1}{1.5} & \textbf{83.1} \ci{83.1}{1.6} & \textbf{84.5} \ci{84.4}{1.4} & 85.3 \ci{85.3}{1.4} & \textbf{83.1} \\ 
\hline
& \multicolumn{8}{c}{\cellcolor{white} \textbf{CPL ($\uparrow$)}} \\
\hline
\rowcolor{gray!20}  Casc. & 81.8 \ci{81.8}{1.9} & 83.4 \ci{83.3}{1.8} & 85.2 \ci{85.2}{1.7} & 81.4 \ci{81.4}{1.9} & 83.3 \ci{83.2}{1.9} & 78.1 \ci{78.1}{2.0} & 53.3 \ci{53.3}{3.0} & 78.1 \\ 
Dir. & \textbf{88.9} \ci{88.9}{1.5} & \textbf{94.0} \ci{94.0}{1.1} & \textbf{91.9} \ci{91.9}{1.2} & \textbf{89.3} \ci{89.2}{1.5} & \textbf{84.0} \ci{84.0}{1.8} & \textbf{88.2} \ci{88.2}{1.5} & \textbf{92.1} \ci{92.1}{1.2} & \textbf{89.8} \\ 
\hline
\hline
& \multicolumn{8}{c}{\cellcolor{white} \textbf{CPS ($\uparrow$)}} \\
\hline
\rowcolor{gray!20} Casc. & \textbf{69.1} \ci{69.1}{2.6} & \textbf{74.0} \ci{73.9}{2.7} & \textbf{64.3} \ci{64.3}{2.9} & \textbf{71.2} \ci{71.2}{2.8} & \textbf{74.4} \ci{74.4}{2.5} & \textbf{74.7} \ci{74.7}{2.6} & \textbf{76.2} \ci{76.2}{2.4} & \textbf{72.0} \\ 
Dir. & 65.4 \ci{65.4}{2.7} & 68.4 \ci{68.3}{2.7} & 60.7 \ci{60.8}{2.8} & 67.9 \ci{67.9}{2.6} & 72.2 \ci{72.2}{2.6} & 71.9 \ci{71.8}{2.7} & 76.0 \ci{75.9}{2.4} & 68.9 \\ 
\toprule
\end{tabular}
}
\caption{Cascade (Casc.) and direct (Dir.) results on all MuST-Cinema language pairs with 95\% CI in parentheses.}
\label{tab:constrained}
\end{table*}

\subsection{Cascade vs. Direct}
\label{subsec:direct_vs_cascade}
After selecting LEV as our best timestamp projection method, we evaluate cascade and direct ST systems trained in the same data condition.
Before this, to ensure the competitiveness of our cascade baseline, we compare it with the results obtained on the MuST-Cinema test set by the other cascade systems presented in literature, namely: en$\rightarrow$\{de, fr\} by \citet{karakanta-etal-2021-flexibility}, and en$\rightarrow$fr by \citet{xu-etal-2022-joint}. As these works report only BLEU with breaks, that is BLEU computed including also \eob~and \eol, we compare our cascade baseline with them on that metric.\footnote{\eob~and \eol~are considered as a single token and replaced, respectively, with § and µ as in the EvalSub toolkit.}
Although these works leverage large additional training corpora for both ASR (e.g. LibriSpeech -- \citealt{librispeech}) and MT (e.g. OPUS -- \citealt{tiedemann-2016-opus} -- and WMT-14 -- \citealt{bojar-etal-2014-findings}), our cascade trained only on MuST-Cinema performs on par with them. 
It scores 20.2 on German and 26.2 on French, which are similar or even better than, respectively, 19.9 and 26.9 of \citep{karakanta-etal-2021-flexibility}, and 25.8 on French of \citep{xu-etal-2022-joint}.
These results confirm the strength of our baselines, and the soundness of our experimental settings.

Table \ref{tab:constrained} reports the scores of the constrained direct and cascade models.
The overall subtitle quality of the direct solution is significantly higher compared to that of the cascade on all language pairs, with a SubER decrease of 3.8-5.5 points, corresponding to an $\sim$8\% improvement on average. 
Since SubER measures translation, segmentation and timestamp quality altogether, to disentangle the contribution of each of these aspects we leverage the other metrics. 
The higher Sigma of our system (+1.2 average improvement) demonstrates that the joint generation of subtitle content and boundaries results in superior segmentation. This finding corroborates previous research on the value of prosody (see \S\ref{subsec:automatic-subtitling}), and the ineffectiveness of projecting caption segmentation onto subtitles, as done by cascade approaches \citep{Georgakopoulou_2019,koponen-etal-2020-mt}. 
The sub-optimal placement of block boundaries in the cascade system can also account for the superior translation quality of our method (+3.9 BLEU average improvement): as the MT component translates the caption block-by-block, inaccurate boundaries can impede access to information required for proper translation.

Looking at the conformity metrics, the direct system complies with the length requirement of 42 characters (CPL) in almost 90\% of cases while the cascade system does so in only 78.1\%.
This difference is explained by the higher number of \eol~generated by the direct model (10-15\% more than the cascade), although being still lower than that of the reference (8-10\% less).
According to the statistics computed on the outputs of the two systems, the cascade does not only have a higher average number of characters per line (32 vs. 29), but its variance is 1.5-2 times greater, with lines sometimes close to or even longer than 100 characters on all language pairs. 
In contrast, most of the CPL violations of the direct system are caused by lines shorter than 60 characters, and lines never exceed 70 characters.
The trend for CPS is instead different, since the cascade generates subtitles with a higher conformity to the 21-CPS reading speed (72.0 vs 68.9). This can be partially explained by looking at the generated timestamps: upon a manual inspection of 100 subtitles, we noticed that the direct model tends to assign the start times of the subtitles slightly after those of the cascade (within 100ms of difference), and end times slightly before those of the cascade (mostly within 200ms). Overall, on the MuST-Cinema test sets, this leads to a total of $\sim$2,940s with subtitles on screen for the cascade, and $\sim$2,850s for the direct ($\sim$3\% lower).

To sum up, our direct system proves to be the best choice to address the automatic subtitling task in the constrained data condition, reaching better translation quality and more well-formed subtitles. Our results also indicate
that improving the reading speed of the generated subtitles is one of the main aspects on which to focus future works.

\begin{table*}[!t]
\centering
\small
\begin{tabular}{l|ccccc} 
\bottomrule
\multicolumn{6}{c}{\textbf{en-de}} \\ 
\hline
\rowcolor{gray!10}
Model & SubER ($\downarrow$) & BLEU~($\uparrow$) & Sigma ($\uparrow$) & CPL ($\uparrow$) & CPS ($\uparrow$) \\ 
\hline
System 1 & 66.9 \ci{66.9}{2.8} & 20.1 \ci{20.2}{1.5} & 71.7 \ci{71.6}{2.4} & \textbf{100} \ci{100}{0.0} & 58.7 \ci{58.6}{3.1} \\
System 2 & 61.5 \ci{61.5}{2.4} & 22.3 \ci{22.2}{1.6} & 71.8 \ci{71.8}{2.3} & \textbf{100} \ci{100}{0.0} & 76.2 \ci{76.2}{2.7} \\
System 3 & 68.1 \ci{68.1}{1.5} & 13.5 \ci{13.5}{1.2} & 62.1 \ci{62.0}{2.6} & 91.6 \ci{91.7}{1.4} & \textbf{89.3} \ci{89.3}{1.8} \\
System 4 & 67.5 \ci{67.1}{7.3} & 23.3 \ci{23.2}{1.7} & 57.9 \ci{57.9}{2.2} & 96.4 \ci{96.4}{0.9} & 83.7 \ci{83.7}{2.3} \\
System 5 & 66.8 \ci{66.8}{2.9} & 19.5 \ci{19.5}{1.5} & 74.0 \ci{74.0}{2.0} & 44.1 \ci{42.8}{3.0} & 50.2 \ci{50.2}{3.1} \\
\hline
Ours & \textbf{59.9} \ci{59.9}{3.2} & \textbf{23.4} \ci{23.4}{1.6} & \textbf{77.9} \ci{78.0}{2.1} & 86.9 \ci{86.9}{1.6} & 68.6 \ci{68.6}{2.7} \\
\specialrule{.1em}{.05em}{.05em} 
\multicolumn{6}{c}{\textbf{en-es}} \\ 
\hline
\rowcolor{gray!10}
Model & SubER ($\downarrow$) & BLEU~($\uparrow$) & Sigma ($\uparrow$) & CPL ($\uparrow$) & CPS ($\uparrow$) \\ 
\hline
System 1 & 52.2 \ci{52.2}{2.7} & 33.4 \ci{33.3}{1.8} & 76.9 \ci{76.9}{1.9} & \textbf{100} \ci{100}{0.0} & 64.6 \ci{64.6}{2.9} \\
System 2 & 51.3 \ci{51.2}{2.4} & 32.7 \ci{32.6}{1.8} & 77.1 \ci{77.0}{2.0} & \textbf{100} \ci{100}{0.0} & 77.6 \ci{77.6}{2.5} \\
System 3 & 58.3 \ci{58.3}{1.7} & 23.3 \ci{23.2}{1.4} & 66.1 \ci{66.0}{2.3} & 94.1 \ci{94.1}{1.2} & \textbf{87.1} \ci{87.1}{2.0} \\
System 4 & 53.8 \ci{53.8}{4.7} & 35.3 \ci{35.3}{2.0} & 65.7 \ci{65.7}{1.8} & 81.3 \ci{81.3}{2.2} & 86.2 \ci{86.1}{2.2} \\
System 5 & 64.6 \ci{64.6}{2.0} & 18.6 \ci{18.6}{1.3} & 79.3 \ci{79.3}{1.9} & 48.5 \ci{48.5}{3.0} & 63.0 \ci{62.9}{2.8} \\
\hline
Ours & \textbf{46.8} \ci{46.7}{2.2} & \textbf{37.4} \ci{37.5}{2.0} & \textbf{81.6} \ci{81.7}{1.5} & 93.2 \ci{93.3}{1.1} & 74.6 \ci{74.6}{2.5} \\
\toprule
\end{tabular}
\caption{Unconstrained results on MuST-Cinema with 95\% CI in  parentheses.}
\label{tab:amara}
\end{table*}

\subsection{Comparison with Production Tools}
\label{subsec:production-results}

To test our approach in more realistic conditions, we train our models on several openly-available corpora (unconstrained condition) and compare them with production tools, which represent very challenging competitors as they can leverage large proprietary datasets. 
We focus on two language pairs (en$\rightarrow$\{de, es\}) for both the in-domain MuST-Cinema, and on the two out-of-domain EC Short Clips and EuroParl Interviews test sets. 
We feed all systems with the full test audio clips, so each system has to segment its audio.
Only in the case of EC Short Clips, and EuroParl Interviews, we clean the audio using Veed\footnote{\url{https://www.veed.io/}.} before processing it, for the sake of a fair comparison with production tools that have similar procedures.\footnote{E.g., see \url{https://www.apptek.com/post/asr-in-captions-accessibility-series-article-7} and \url{https://sonix.ai/articles/how-to-remove-background-audio-noise}.} The impact of audio cleaning is analyzed in Appendix \ref{app:noise}.

\paragraph{MuST-Cinema}
The results of the unconstrained models on the in-domain MuST-Cinema test set are shown in Table \ref{tab:amara}. 
Compared to production tools, our system shows better translation and segmentation quality as well as a significantly better overall quality on both languages.
Gains in BLEU are more evident in Spanish, where we obtain a $\sim$6\% improvement compared to the second-best model (System 4). 
Also, considerable Sigma improvements are observed with gains of 5.3-34.5\% for German and 2.9-24.2\% for Spanish, which are in line with SubER improvements of, respectively, 2.6-12.0\% and 8.8-27.6\%. A perfect CPL conformity is reached by System 1 and 2 for both languages, while our system is on par with System 3 on en-es and falls slightly behind System 3 and 4 on en-de, with a $\sim$90\% average conformity for the two language pairs.
System 5 is by far the worst, as it violates the 42 CPL constraint in more than 50\% of the lines.
As for CPS conformity, we observe that our system achieves better scores compared to System 1 and 5 but it is worse than System 2, 3, and 4 on both language directions, highlighting again the need to improve this aspect in future work.

\begin{table*}[!t]
\centering
\small
\begin{tabular}{l|ccccc} 
\bottomrule
\multicolumn{6}{c}{\textbf{en-de}} \\ 
\hline
\rowcolor{gray!10}
Model & SubER ($\downarrow$) & BLEU~($\uparrow$) & Sigma ($\uparrow$) & CPL ($\uparrow$) & CPS ($\uparrow$) \\ 
\hline
System 1 & 63.0 \ci{63.0}{2.4} & 23.8 \ci{23.8}{1.9} & \textbf{71.6} \ci{71.5}{2.7} & \textbf{100} \ci{100}{0.0} & 76.1 \ci{76.1}{2.8} \\
System 2 & 60.8 \ci{60.8}{1.8} & 22.1 \ci{22.1}{1.9} & 67.2 \ci{67.1}{2.9} & \textbf{100} \ci{100}{0.0} & 91.1 \ci{91.1}{1.9} \\
System 3 & \textbf{59.0} \ci{58.9}{1.9} & 25.0 \ci{25.0}{1.9} & 70.4 \ci{70.4}{2.8} & 84.6 \ci{84.6}{1.9} & \textbf{95.4} \ci{95.4}{1.4} \\
System 4 & 61.5 \ci{61.5}{3.3} & \textbf{28.2} \ci{28.3}{2.0} & 59.4 \ci{59.4}{2.2} & \textbf{100} \ci{100}{0.0} & 94.9 \ci{95.0}{1.5} \\
System 5 & 62.4 \ci{62.4}{2.2} & 24.2 \ci{24.2}{1.8} & 71.3 \ci{71.2}{2.2} & 39.8 \ci{39.7}{3.4} & 71.3 \ci{71.3}{3.3} \\
\hline
Ours & 59.9 \ci{59.9}{2.2} & 25.3 \ci{25.3}{1.9} & 70.8 \ci{70.7}{2.4} & 81.3 \ci{81.3}{2.2} & 79.9 \ci{80.0}{2.7} \\
\specialrule{.1em}{.05em}{.05em} 
\multicolumn{6}{c}{\textbf{en-es}} \\ 
\hline
\rowcolor{gray!10}
Model & SubER ($\downarrow$) & BLEU~($\uparrow$) & Sigma ($\uparrow$) & CPL ($\uparrow$) & CPS ($\uparrow$) \\ 
\hline
System 1 & 52.9 \ci{52.9}{1.8} & 33.7 \ci{33.7}{1.8} & 76.0 \ci{75.9}{2.2} & \textbf{100} \ci{100}{0.0} & 80.4 \ci{80.3}{2.8} \\
System 2 & 51.7 \ci{51.6}{1.6} & 32.2 \ci{32.3}{1.9} & 75.6 \ci{75.6}{2.2} & \textbf{100} \ci{100}{0.0} & 93.5 \ci{93.5}{1.7} \\
System 3 & \textbf{49.7} \ci{49.7}{1.8} & 35.5 \ci{35.5}{1.8} & 74.9 \ci{74.9}{1.9} & 87.3 \ci{87.4}{1.8} & \textbf{95.3} \ci{95.3}{1.4} \\
System 4 & 50.2 \ci{50.2}{2.2} & \textbf{39.6} \ci{39.6}{1.9} & 61.9 \ci{61.9}{1.8} & \textbf{100} \ci{100}{0.0} & 93.4 \ci{93.4}{1.4} \\
System 5 & 64.9 \ci{64.9}{1.6} & 21.9 \ci{21.9}{1.5} & \textbf{79.7} \ci{79.6}{2.0} & 41.7 \ci{41.6}{3.3} & 73.1 \ci{73.0}{3.2} \\
\hline
Ours & 52.7 \ci{52.7}{2.0} & 34.8 \ci{34.9}{2.0} & 72.6 \ci{72.7}{2.0} & 88.6 \ci{88.5}{1.6} & 79.1 \ci{79.0}{2.6} \\
\toprule
\end{tabular}
\caption{Unconstrained results on EC Short Clips with 95\% CI in parentheses.}
\label{tab:documentary}
\end{table*}

\paragraph{EC Short Clips}
This out-of-domain test set presents additional difficulties compared to TED talks, namely the presence of multiple speakers and background music during speech.
It is worth mentioning that our direct ST models have not been trained to be robust to these phenomena, as they are not present in the training data, whereas production tools are designed to deal with any condition,
and may have dedicated modules to handle them.

Nevertheless, the results in Table \ref{tab:documentary} show that, even in these challenging conditions, our direct ST models are competitive with production tools on BLEU, Sigma, and SubER.
Indeed, there is no clear winner between the systems as the best score for each metric is obtained by a different model, which also varies across languages.
Looking at the conformity constraints, Systems 1, 2, and 4 achieve a perfect CPL conformity (100\%), while ours is comparable with System 3 and better than System 5.
This difference is likely motivated by the number of \eol~inserted by our system, which is considerably lower than that of System 4 (368 vs. 635 for German and 451 vs. 594 for Spanish).  
Instead, the results for CPS conformity follow the same trend observed in the constrained data condition (§\ref{subsec:direct_vs_cascade}).

\begin{table*}[t]
\centering
\small
\begin{tabular}{l|ccccc} 
\bottomrule
\multicolumn{6}{c}{\textbf{en-de}} \\ 
\hline
\rowcolor{gray!10}
Model & SubER ($\downarrow$) & BLEU~($\uparrow$) & Sigma ($\uparrow$) & CPL ($\uparrow$) & CPS ($\uparrow$) \\ 
\hline
System 1 & 84.9 \ci{85.0}{2.4} & 12.3 \ci{12.3}{1.1} & 64.8 \ci{64.8}{2.8} & \textbf{100} \ci{100}{0.0} & 67.6 \ci{67.7}{2.8} \\
System 2 & 78.4 \ci{78.4}{2.0} & 13.2 \ci{13.2}{1.1} & 63.9 \ci{63.9}{2.9} & \textbf{100} \ci{100}{0.0} & 79.8 \ci{79.8}{2.3} \\
System 3 & \textbf{78.1} \ci{78.1}{1.9} & 13.6 \ci{13.6}{1.1} & 69.6 \ci{69.6}{2.8} & 86.9 \ci{86.9}{1.6} & \textbf{93.2} \ci{93.3}{1.4} \\
System 4 & 80.1 \ci{80.1}{2.7} & \textbf{15.8} \ci{15.8}{1.3} & 56.9 \ci{56.9}{2.8} & \textbf{100} \ci{100}{0.0} & 83.8 \ci{83.9}{2.2} \\
System 5 & 85.1 \ci{85.1}{1.9} & 11.4 \ci{11.4}{1.1} & 69.8 \ci{69.8}{2.5} & 44.4 \ci{44.4}{2.8} & 59.2 \ci{59.3}{2.7} \\
\hline
Ours & 80.3 \ci{80.3}{2.4} & 12.5 \ci{12.5}{1.1} & \textbf{70.0} \ci{70.0}{2.8} & 80.9 \ci{81.0}{1.9} & 68.8 \ci{68.8}{2.5} \\
\specialrule{.1em}{.05em}{.05em} 
\multicolumn{6}{c}{\textbf{en-es}} \\ 
\hline
\rowcolor{gray!10}
Model & SubER ($\downarrow$) & BLEU~($\uparrow$) & Sigma ($\uparrow$) & CPL ($\uparrow$) & CPS ($\uparrow$) \\ 
\hline
System 1 & 75.5 \ci{75.5}{2.3} & 19.8 \ci{19.8}{1.3} & 72.7 \ci{72.7}{2.2} & \textbf{100} \ci{100}{0.0} & 72.7 \ci{72.8}{2.5} \\
System 2 & 71.4 \ci{71.4}{2.1} & 20.9 \ci{20.9}{1.4} & 73.8 \ci{73.8}{2.0} & \textbf{100} \ci{100}{0.0} & 81.4 \ci{81.5}{2.3} \\
System 3 & 70.0 \ci{70.1}{2.2} & 20.8 \ci{20.8}{1.4} & 72.8 \ci{72.8}{2.0} & 90.5 \ci{90.5}{1.4} & \textbf{93.7} \ci{93.7}{1.3} \\
System 4 & \textbf{68.6} \ci{68.5}{2.5} & \textbf{25.4} \ci{25.4}{1.4} & 61.6 \ci{61.6}{2.0} & \textbf{100} \ci{100}{0.0} & 91.5 \ci{91.5}{1.8} \\
System 5 & 80.8 \ci{80.8}{1.7} & 13.0 \ci{12.9}{1.1} & \textbf{77.3} \ci{77.3}{2.4} & 52.1 \ci{52.1}{2.8} & 67.4 \ci{67.5}{2.7} \\
\hline
Ours & 72.3 \ci{72.3}{2.2} & 20.8 \ci{20.9}{1.4} & 70.4 \ci{70.4}{2.0} & 90.1 \ci{90.1}{1.3} & 76.9 \ci{76.9}{2.4} \\
\toprule
\end{tabular}
\caption{Unconstrained results on EuroParl Interviews with 95\% CI in parentheses.}
\label{tab:interviews}
\end{table*}

Even though this scenario features completely different domain and audio characteristics, some trends are in line with the results shown in Table \ref{tab:amara}. 
System 3 always achieves the best CPS conformity, while Systems 1, 2, and 4 achieve perfect CPL conformity on both languages.
Moreover, although System 4 achieves the best translation quality (and it is the second best on MuST-Cinema, after our system), its segmentation quality (Sigma) is always the worst, indicating that its subtitles are not segmented in an optimal way to facilitate comprehension. 
All in all, these results suggest that each production tool has been optimized on a different aspect of automatic subtitling (e.g. System 3 has been optimized to achieve high CPS conformity). In contrast, our direct model, which has been trained without prioritizing any specific aspect, performs on average, also achieving competitive results in out-of-domain scenarios.

\begin{table*}[!tb]
\centering
\small
\setlength{\tabcolsep}{4pt}
\begin{tabular}{l||c|c|c|c|c|c} 
 & System 1 & System 2 & System 3 & System 4 & System 5 & Ours \\ 
\hline
en-de & 72.0 \ci{72.0}{1.6} & 67.2 \ci{67.1}{1.3} & 69.0 \ci{69.0}{1.2} & 70.1 \ci{70.1}{3.3} & 71.9 \ci{71.9}{1.7} & \textbf{67.0} \ci{67.0}{1.7} \\
\hline
en-es & 60.3 \ci{60.3}{1.5} & 58.2 \ci{58.2}{1.3} & 59.8 \ci{59.8}{1.2} & 57.8 \ci{57.8}{2.4} & 70.2 \ci{70.2}{1.1} & \textbf{57.2} \ci{57.1}{1.5} \\
\toprule
\end{tabular}
\caption{SubER ($\downarrow$) over the three test sets with 95\% CI in parentheses.}
\label{tab:final}
\end{table*}

\paragraph{EuroParl Interviews} 
EuroParl Interviews represents the most difficult of the three test sets: it contains multiple speakers, and the target translations are not verbatim since they are compressed to perfectly fit the subtitling constraints (§\ref{sec:guidelines}). This characteristic is very challenging for current automatic subtitling tools, especially for our direct model since it has not been trained on similar data.

The results are shown in Table \ref{tab:interviews}. As on the EC test set, our system performs competitively with production tools, even achieving the best Sigma for German.
For CPL, instead, most systems have high length conformity, even reaching 100\%. 
As already noticed on the other test sets, the CPL conformity is strongly correlated with the number of \eol~inserted by a system: our model has an average conformity of 85.5\% with only 451~\eol~inserted, nearly half of those inserted by System~1~(864), System~2~(711), and System~4~(774) that always comply with the CPL constraint.
CPS conformity shows the same trend as with the other test sets.

Compared to the results in Tables \ref{tab:amara} and \ref{tab:documentary}, we can see that all systems struggle in achieving a comparable overall subtitle quality (SubER), high-quality segmentations (Sigma), and, above all, high translation quality (BLEU). 
The translation quality of all systems degrades by at least 10 BLEU compared to 
the values observed on the MuST-Cinema and EC test sets. However, as previously mentioned, these results are expected since the EuroParl Interviews test set contains condensed translations of the source speech.

All in all, we can conclude that our direct ST model, even though not developed as a production-ready system (it is not trained on huge amounts of data and different domains), is competitive with production tools.
Indeed, considering the SubER metric computed over the three test sets (Table \ref{tab:final}), our direct ST approach is the best on both German (67.0) and Spanish (57.2).
As only the scores of System 2 fall within the confidence interval of our direct model in both cases, we can conclude that our model is on par with the best production system and outperforms the others in terms of SubER.

\section{Conclusions}
In this paper, we proposed the first approach based on direct speech-to-text translation models to fully automatize the subtitling process, including translation, segmentation into subtitles, and timestamp estimation.
Experiments in constrained data conditions on 7 language pairs demonstrated the potential of our approach, which outperformed the current cascade architectures with a $\sim$7\% improvement in terms of SubER.
In addition, to test the generalisability of our findings across subtitling genres, we extended our evaluation setting by collecting two new test sets for en$\rightarrow$\{de, es\} covering different domains, degrees of subtitle condensation, and audio conditions.
Finally, we compared our models with production tools in unconstrained data conditions on both existing benchmarks and the newly-collected test sets.
This comparison further highlighted that our approach represents a promising direction: although trained on a relatively limited amount of data, our systems achieved comparable quality with production tools, with improvements in SubER ranging from 0.2 to 5.0 on en$\rightarrow$de and from 0.6 to 13.0 on en$\rightarrow$es over the three test sets.

\appendix

\section{Timestamp Extraction Method}
\label{sec:force-align}

To validate the effectiveness of extracting source-side timestamps with the CTC-based segmentation algorithm, we conduct an ablation study, where we replace it with the forced aligner tool of the Cascade architecture (\S\ref{subsec:comparison}).
Table \ref{tab:force-align} reports the scores. The forced aligner tool (FA) achieves similar results compared to the CTC-based segmentation algorithm (CTC), with a slightly worse SubER (+0.1) on average on the three test sets. 
Moreover, it is important to highlight that our method does not require an external model. 
These findings support our choice and align with previous research by \citet{ctcsegmentation}, which highlighted the competitiveness of the CTC-based segmentation approach compared to widely used forced aligners (in their case, Gentle\footnote{\url{https://github.com/lowerquality/gentle}}).

\begin{table}[!htb]
\centering
\small
\setlength{\tabcolsep}{2.65pt}
\begin{tabular}{l|ccc|ccc|c}
\bottomrule
\multirow{2}{*}{\textbf{Method}} & \multicolumn{3}{c|}{\textbf{en-de}} & \multicolumn{3}{c|}{\textbf{en-es}} & \multirow{2}{*}{\textbf{Avg.}} \\
\cline{2-7}
 & \textbf{MC} & \textbf{ECSC} & \textbf{EPI} & \textbf{MC} & \textbf{ECSC} & \textbf{EPI} & \\
\hline
CTC & 59.9 & \textbf{59.9} & \textbf{80.3} & 46.8 & \textbf{52.7} & 72.3 & \textbf{62.0} \\
FA & \textbf{59.7} & 60.3 & 80.7 & \textbf{46.7} & \textbf{52.7} & \textbf{72.2} & 62.1 \\
\toprule
\end{tabular}
\caption{\label{tab:force-align} SubER scores ($\downarrow$) on MuST-Cinema test set (MC), EC Short Clips (ECSC), and EuroParl Interviews (EPI) when the CTC-based audio segmentation (CTC) or the forced aligner (FA) method is used to extract the source-side timestamps.}
\end{table}

\section{Effect of Background Noise}
\label{app:noise}
The presence of background noise in the test sets complicates both the audio segmentation (performed with SHAS) and the generation with the direct ST model. For this reason, for the sake of a fair comparison with production tools, we used Veed to remove the background noise from EC Short Clips and EuroParl Interviews, as mentioned in \S\ref{subsec:production-results}.
Table \ref{tab:noise} shows the impact of background noise on the resulting subtitling quality. By comparing \textit{1.} and \textit{3.}, we notice that the presence of background noise causes an overall relative error increase of $\sim$5\% on average over the two test sets and two language pairs. 
The degradation is caused both by the lower quality of the audio segmentation of SHAS and by worse outputs produced by the direct ST system, as the absence of noise during segmentation (\textit{2.}) improves by an average of 1.7 SubER the results obtained without noise removal (\textit{3.}).
Creating models robust to background noise, though, is a task \textit{per se} \citep{Seltzer-2013-nouse,Jinyu-2014-noise-robust,Mitra2017} and goes beyond the scope of this work.

\begin{table}[!htb]
\centering
\small
\setlength{\tabcolsep}{4.25pt}
\begin{tabular}{l|cc|cc|c}
\bottomrule
\textbf{Noise}& \multicolumn{2}{c|}{\textbf{en-de}} & \multicolumn{2}{c|}{\textbf{en-es}} & \multirow{2}{*}{\textbf{Avg.}}           \\
\cline{2-5}
\textbf{Removed} & \textbf{ECSC} & \textbf{EPI} & \textbf{ECSC} & \textbf{EPI} &  \\
\hline
1. Yes & 59.9 & 80.3 & 66.3 & 72.3 & 52.7 \\
2. Only Segm. & 61.4 & 82.0 & 68.4 & 73.9 & 56.4 \\
3. No & 63.1 & 81.7 & 69.5 & 75.3 & 58.1 \\
\toprule
\end{tabular}
\caption{SubER scores ($\downarrow$) on EC Short Clips (ECSC) and EuroParl Interviews (EPI) with background noise removal for: both the audio segmentation with SHAS and the prediction of the direct ST system (\textit{1.}); only the audio segmentation, but the noisy audio is fed as input to the direct ST model (\textit{2.}); no noise removal (\textit{3.}).}
\label{tab:noise}
\end{table}

\section*{Acknowledgements}
We acknowledge the support  of the PNRR project FAIR - Future AI Research (PE00000013),  under the NRRP MUR program funded by the NextGenerationEU.
The authors thank Marco Matassoni (and the SpeechTek unit at FBK) for their help in providing forced aligners, and Evgeny Matusov (and AppTek) for sharing their production system outputs, as well as the anonymous reviewers for their insightful comments that improved the manuscript.

\bibliography{tacl2021}
\bibliographystyle{acl_natbib}

\end{document}